# Learning causation event conjunction sequences

Thomas E. Portegys, portegys@gmail.com ORCID 0000-0003-0087-6363

Dialectek, DeKalb, Illinois USA

## Abstract

This is an examination of some methods that learn causations in event sequences. A causation is defined as a conjunction of one or more cause events occurring in an arbitrary order, with possible intervening non-causal events, that lead to an effect. The methods include recurrent and non-recurrent artificial neural networks (ANNs), as well as a histogram-based algorithm. An attention recurrent ANN performed the best of the ANNs, while the histogram algorithm was significantly superior to all the ANNs.

## Introduction

This is an examination of some methods that learn causations in event sequences. A causation is defined as a conjunction of one or more cause events occurring in an arbitrary order, with possible intervening non-causal events, that lead to an effect. A claim is made that situations in the world call for this type of learning. The methods include recurrent and non-recurrent artificial neural networks (ANNs), as well as a histogram-based algorithm.

When a bird builds a nest, it needs to know the current structure state in order to add on the next component. It can be seen flitting about here and there while assessing things, the wind blowing the nest around. When a person follows a recipe to prepare food, the macro-instructions are in sequence, but the chef's detailed movements are often in zig-zag fashion. These might be thought of as quasi-sequential tasks, affording many event orderings to accomplish the same end. We actually produce these disorderings: our eyes naturally drift around scenes in Brownian-like trajectories (Poletti et al., 2015) producing streams of visual events that the brain uses to compose perceptual models. Interruptions and digressions also occur in the real world, producing intervening events: the phone rings, a spoon is dropped, or a window needs shutting during a task.

Instead of learning an explosive number of alternative causal event orderings that produce the same effect, encapsulating these events within an unordered set is a more efficient scheme. A *causation* is defined to be such a set. A causation is a conjunction of one or more *cause* events occurring in an arbitrary order. Conjunction means that all cause events must occur to satisfy a causation. Recognizing a causation allows an agent to predict or implement an effect of some sort.

From Aristotle's (384–322 BC) Four Modes, to Pearl's Ladder (2009), causation (also called causality) has been a subject of study for philosophers, mathematicians, psychologists, and others. This is unsurprising, as determining causation is a vital business of life for humans and many animals, allowing them to predict and manipulate their environment. It is especially important for scientific experimentation, where teasing apart causation from correlation can be a statistical challenge.

How do humans "natively" process causation? Novick and Cheng (2004) propose that interacting, or *conjunctive*, causes form independent mental entities. For example, when the conjunctive causes *XY* are presented with a positive effect, and *X* and *Y* are presented alone with negative effects, subjects will infer the existence of a conjunctive entity "*XY*" and will assume that it is a distinct cause independent of *X* and *Y*. Liljeholm (2015) asserts that a Bayesian framework can be used to model this.

Closer to home in the realm of ANNs, Vinyals et al. (2016) discusses the problems that ANNs can have with inputs that are disordered, the classic case being the problem of sorting sets of numbers. While the input may have many orderings, more than training can accommodate, the output is in a fixed order. They introduce a modestly successful model based on a Neural Turing Machine (Graves et al., 2014). A Neural Turing Machine enhances an ANN with an external associative memory that can retrieve input information based on content, thus making it order-independent. In contrast, in this project input events belong to limited-size domains rather than, for example, the domain of numbers. In addition the question of how intervening non-causal events are handled is also posed.

This project was inspired by Mona (Portegys, 2001), a goal-seeking ANN that composes hierarchies of cause-and-effect neurons. A method for representing order-independent causations would be a valuable efficiency enhancement.

# Description

The code and instructions are available at https://github.com/portegys/CausationLearning.

## Causations and instances

Causations are randomly generated from the parameters shown in Table 1.

| Parameter | Description |
|---|---|
| NUM_EVENT_TYPES | Number of event types. |
| MAX_CAUSE_EVENTS | Maximum cause events per instance. |
| MAX_INTERVENING_EVENTS | In a valid causation, the maximum number of non-causal events between causal events. |
| NUM_CAUSATIONS | Number of causations. |

Table 1 – Causation parameters.

In Figure 1 there are two causations generated from 10 possible event types. Causation 0 consists of a single event: 4. Causation 1 specifies that event 2 must appear at least twice in an instance.

```
Causations:
[0] ID=0, events: { 4 }
[1] ID=1, events: { 2 2 }

Causation instances:
[0] Events: { 4 9 9 3 0 } Causation IDs: { 0 }
[1] Events: { 4 2 2 8 0 } Causation IDs: { 0 1 }
[2] Events: { 3 2 4 2 8 } Causation IDs: { 0 1 }
[3] Events: { 6 2 2 7 9 } Causation IDs: { 1 }
[4] Events: { 2 9 2 1 6 } Causation IDs: { 1 }
[5] Events: { 8 8 0 2 3 } Causation IDs: { }
[6] Events: { 5 1 7 8 1 } Causation IDs: { }
```

```
[7] Events: { 8 3 0 5 2 } Causation IDs: { }
[8] Events: { 0 5 9 8 2 } Causation IDs: { }
[9] Events: { 3 2 9 8 6 } Causation IDs: { }
```

Figure 1 – Sample causations and causation instances.

A causation instance is a sequence of events that contain 0 or more causations. To create a causation instance, a random stream of event types is generated. The stream is then checked to see if causations appear in the stream. A valid instance contains at least one causation. As shown in Figure 1, instance 0 contains causation 0, indicated by the red event type 4. Instance 1 contains both causation 0 and 1, shown by the event types in red and blue. Note how causation 1's events are separated by other events in instance 4. If cause events are separated by a number of events exceeding the MAX_INTERVENING_EVENTS parameter, the instance is also invalid.

### ANN specifications

The Keras 2.6.0 python machine learning system is used for all the networks. Input is one-hot encoded. Output is causation iDs multiple-hot encoded, as multiple causation IDs are possible in an instance. Training is conducted with 500 epochs, resulting in final losses of around 0.005. The recurrent ANNs were an LSTM (Long Short-Term Memory) (Hochreiter, S., Schmidhuber, J., 1997) and an LSTM with an Attention layer (Vaswani et al., 2017). . Attention networks are noted for their ability to detect latent patterns in a sequences, such as found in language. A non-recurrent ANN was also evaluated. These are the specifications for each network:

LSTM: 128 neurons in a hidden layer, mean squared error loss function, and Adam optimizer.

Attention: 128 neuron LSTM layer overlaid with 64 neuron attention layer, mean absolute error loss function, Adam optimizer.

NN (non-recurrent): 128 neuron hidden layer, binary cross-entropy loss function, Adam optimizer. Input: each possible event type encoded with steps in which event occurred, if any.

### Histogram algorithm

The algorithm is generally based on methods for statistically detecting signals in noise (Pini, 2019). For each valid training instance, counts of event types are kept for each causation ID. Each valid instance for a causation must contain its relevant event types, while the remaining events are random. With enough instances, the relevant event types will exclusively appear and thus be detectable.

The maximum intervening events is determined either by observing valid training types and measuring the maximum distance between cause events. Again, with enough random instances, this technique will settle on the correct value.

## Results

The three ANNs (LSTM, Attention, and NN), along with the histogram algorithm were trained and tested with varying parameter values shown in Table 2. Each was done with a training set of 100 instances, 50 valid and 50 invalid. Then tested with a set of 50 random instances.

| Parameter | Abbreviation | Description | Values |
|---|---|---|---|
| NUM_EVENT_TYPES | NET | Number of event types. | 15, 20, 25 |
| NUM_CAUSATIONS | NC | Number of causations. | 2, 4, 6, 8, 10 |
| MAX_CAUSE_EVENTS | MCE | Maximum number of cause events. | 1, 2, 3 |
| MAX_INTERVENING_EVENTS | MIE | Maximum number of intervening events | 0, 1, 2 |

Table 2 – Training and testing parameters.

10 runs with different random seeds produced accumulated results. The accuracies were scored with the following scale: *Good*: 100%-90%, *Fair*: 89%-70%, *Poor*: 69%-0%. The overall comparative results are shown in Figure 2.

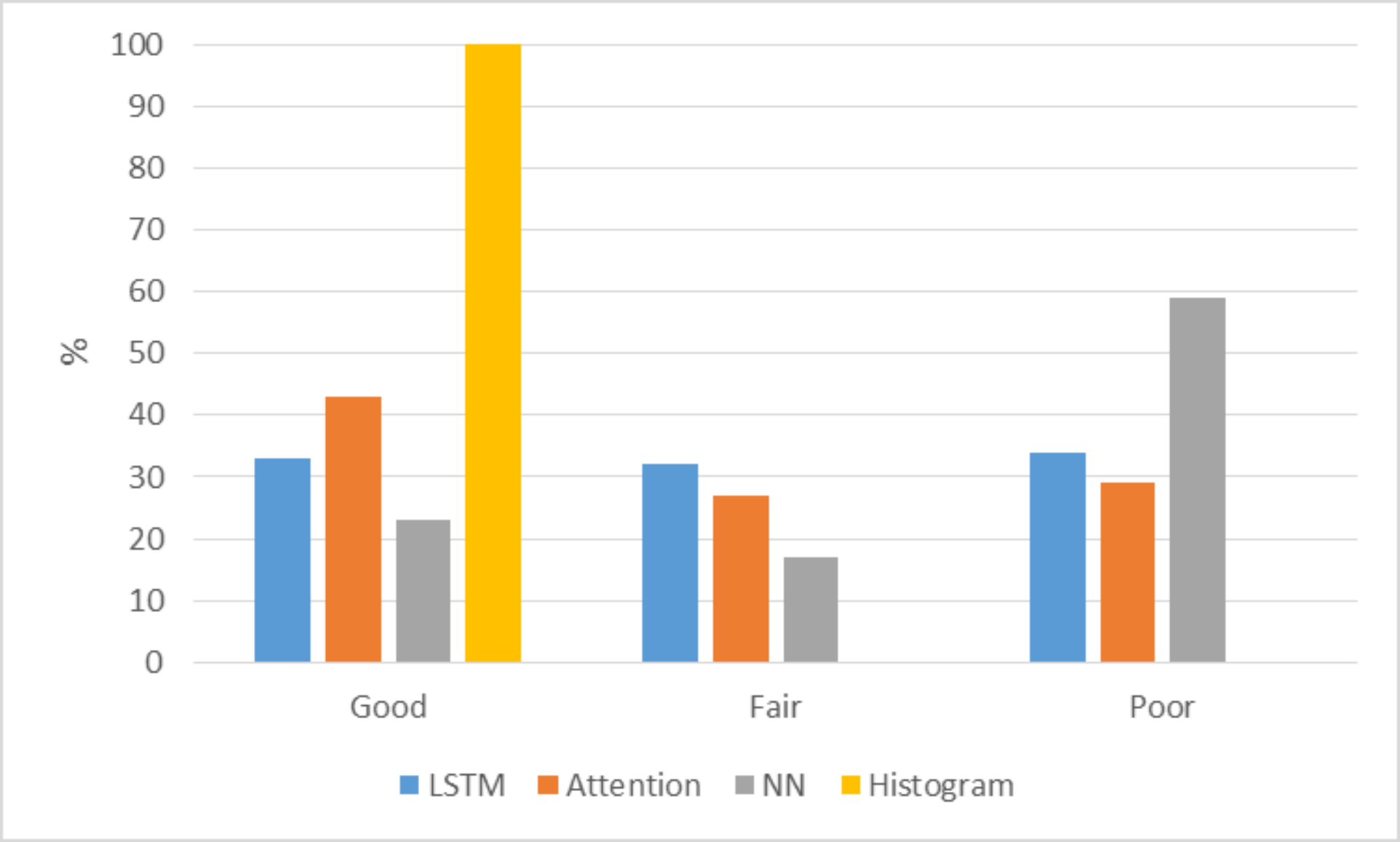

Figure 2 – Overall comparative results.

A breakdown of the results for each approach follows, including a graphical and text decision tree representation. The decision tree displays how the parameter settings determine the scores. Each node indicates a parameter value constraint, number of instances, vector of counts of each class (*Good, Fair, Poor*), and an overall class denoted by the maximum count. Brown nodes are *Good*, green are *Fair*, and blue are *Poor*. The color intensity indicates the density of the classification.

## LSTM

Scores: *Good*=26, *Fair*=27, *Poor*=28. The results are shown in Figures 3 and 4.

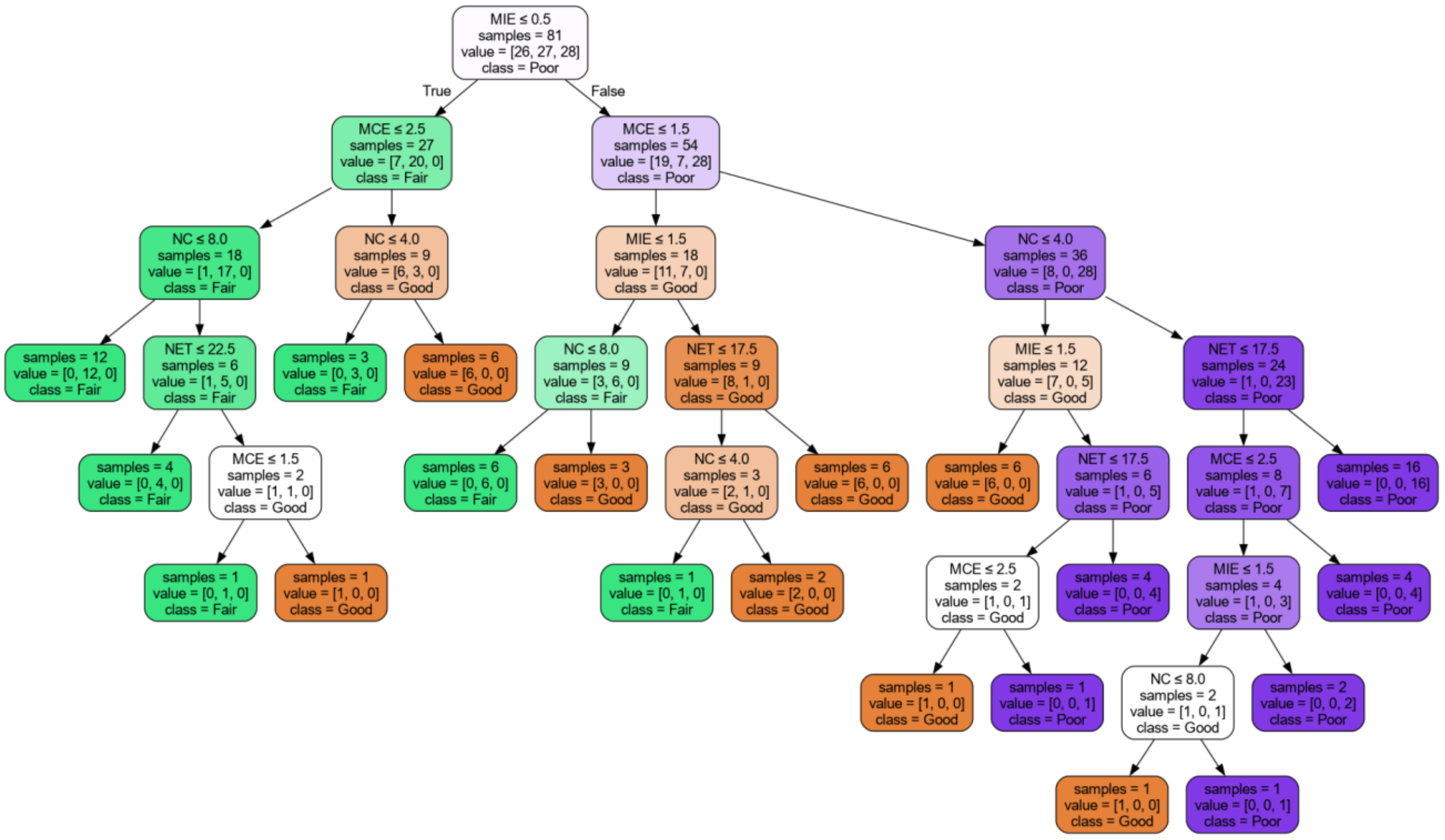

Figure 3 – LSTM graphical decision tree.

```
|--- MAX_INTERVENING_EVENTS <= 0.50
|   |--- MAX_CAUSE_EVENTS <= 2.50
|   |   |--- NUM_CAUSATIONS <= 8.00
|   |   |   |--- class: Good
|   |   |--- NUM_CAUSATIONS >  8.00
|   |   |   |--- NUM_EVENT_TYPES <= 22.50
|   |   |   |   |--- class: Good
|   |   |   |--- NUM_EVENT_TYPES >  22.50
|   |   |   |   |--- MAX_CAUSE_EVENTS <= 1.50
|   |   |   |   |   |--- class: Good
|   |   |   |   |--- MAX_CAUSE_EVENTS >  1.50
|   |   |   |   |   |--- class: Fair
|   |--- MAX_CAUSE_EVENTS >  2.50
|   |   |--- NUM_CAUSATIONS <= 4.00
|   |   |   |--- class: Good
|   |   |--- NUM_CAUSATIONS >  4.00
|   |   |   |--- class: Fair
|--- MAX_INTERVENING_EVENTS >  0.50
|   |--- MAX_CAUSE_EVENTS <= 1.50
|   |   |--- MAX_INTERVENING_EVENTS <= 1.50
|   |   |   |--- NUM_CAUSATIONS <= 8.00
|   |   |   |   |--- class: Good
```

```
|   |   |   |--- NUM_CAUSATIONS >  8.00
|   |   |   |   |--- class: Fair
|   |   |--- MAX_INTERVENING_EVENTS >  1.50
|   |   |   |--- NUM_EVENT_TYPES <= 17.50
|   |   |   |   |--- NUM_CAUSATIONS <= 4.00
|   |   |   |   |   |--- class: Good
|   |   |   |   |--- NUM_CAUSATIONS >  4.00
|   |   |   |   |   |--- class: Fair
|   |   |   |--- NUM_EVENT_TYPES >  17.50
|   |   |   |   |--- class: Fair
|   |--- MAX_CAUSE_EVENTS >  1.50
|   |   |--- NUM_CAUSATIONS <= 4.00
|   |   |   |--- MAX_INTERVENING_EVENTS <= 1.50
|   |   |   |   |--- class: Fair
|   |   |   |--- MAX_INTERVENING_EVENTS >  1.50
|   |   |   |   |--- NUM_EVENT_TYPES <= 17.50
|   |   |   |   |   |--- MAX_CAUSE_EVENTS <= 2.50
|   |   |   |   |   |   |--- class: Fair
|   |   |   |   |   |--- MAX_CAUSE_EVENTS >  2.50
|   |   |   |   |   |   |--- class: Poor
|   |   |   |   |--- NUM_EVENT_TYPES >  17.50
|   |   |   |   |   |--- class: Poor
|   |   |--- NUM_CAUSATIONS >  4.00
|   |   |   |--- NUM_EVENT_TYPES <= 17.50
|   |   |   |   |--- MAX_CAUSE_EVENTS <= 2.50
|   |   |   |   |   |--- MAX_INTERVENING_EVENTS <= 1.50
|   |   |   |   |   |   |--- NUM_CAUSATIONS <= 8.00
|   |   |   |   |   |   |   |--- class: Fair
|   |   |   |   |   |   |--- NUM_CAUSATIONS >  8.00
|   |   |   |   |   |   |   |--- class: Poor
|   |   |   |   |   |--- MAX_INTERVENING_EVENTS >  1.50
|   |   |   |   |   |   |--- class: Poor
|   |   |   |   |--- MAX_CAUSE_EVENTS >  2.50
|   |   |   |   |   |--- class: Poor
|   |   |   |--- NUM_EVENT_TYPES >  17.50
|   |   |   |   |--- class: Poor
```

Figure 4 – LSTM text decision tree.

When there are no intervening events the results significantly improve. Secondarily, when there is a single cause event the results are better. Thirdly, the number of causations has an effect: 4 or less is better.

## Attention

Scores: *Good*=22, *Fair*=35, *Poor*=24. The results are shown in Figures 5 and 6.

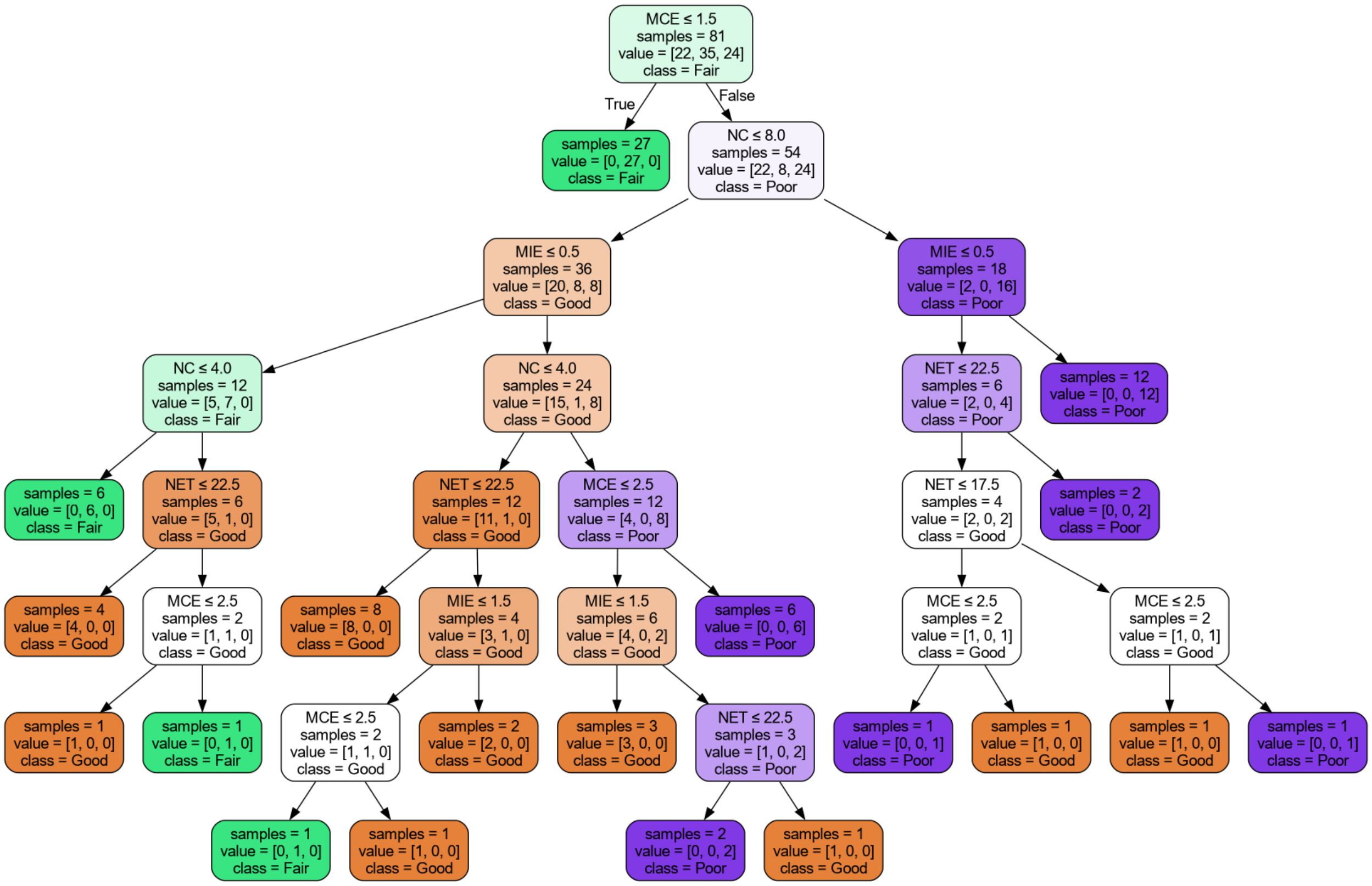

Figure 5 – Attention graphical decision tree.

```
|--- MAX_CAUSE_EVENTS <= 1.50
|   |--- class: Good
|--- MAX_CAUSE_EVENTS >  1.50
|   |--- NUM_CAUSATIONS <= 8.00
|   |   |--- MAX_INTERVENING_EVENTS <= 0.50
|   |   |   |--- NUM_CAUSATIONS <= 4.00
|   |   |   |   |--- class: Good
|   |   |   |--- NUM_CAUSATIONS >  4.00
|   |   |   |   |--- NUM_EVENT_TYPES <= 22.50
|   |   |   |   |   |--- class: Fair
|   |   |   |   |--- NUM_EVENT_TYPES >  22.50
|   |   |   |   |   |--- MAX_CAUSE_EVENTS <= 2.50
|   |   |   |   |   |   |--- class: Fair
|   |   |   |   |   |--- MAX_CAUSE_EVENTS >  2.50
|   |   |   |   |   |   |--- class: Good
|   |   |--- MAX_INTERVENING_EVENTS >  0.50
|   |   |   |--- NUM_CAUSATIONS <= 4.00
|   |   |   |   |--- NUM_EVENT_TYPES <= 22.50
|   |   |   |   |   |--- class: Fair
|   |   |   |   |--- NUM_EVENT_TYPES >  22.50
|   |   |   |   |   |--- MAX_INTERVENING_EVENTS <= 1.50
|   |   |   |   |   |   |--- MAX_CAUSE_EVENTS <= 2.50
```

```
|   |   |   |   |   |   |   |--- class: Good
|   |   |   |   |   |   |--- MAX_CAUSE_EVENTS >  2.50
|   |   |   |   |   |   |   |--- class: Fair
|   |   |   |   |   |--- MAX_INTERVENING_EVENTS >  1.50
|   |   |   |   |   |   |--- class: Fair
|   |   |   |--- NUM_CAUSATIONS >  4.00
|   |   |   |   |--- MAX_CAUSE_EVENTS <= 2.50
|   |   |   |   |   |--- MAX_INTERVENING_EVENTS <= 1.50
|   |   |   |   |   |   |--- class: Fair
|   |   |   |   |   |--- MAX_INTERVENING_EVENTS >  1.50
|   |   |   |   |   |   |--- NUM_EVENT_TYPES <= 22.50
|   |   |   |   |   |   |   |--- class: Poor
|   |   |   |   |   |   |--- NUM_EVENT_TYPES >  22.50
|   |   |   |   |   |   |   |--- class: Fair
|   |   |   |   |--- MAX_CAUSE_EVENTS >  2.50
|   |   |   |   |   |--- class: Poor
|   |--- NUM_CAUSATIONS >  8.00
|   |   |--- MAX_INTERVENING_EVENTS <= 0.50
|   |   |   |--- NUM_EVENT_TYPES <= 22.50
|   |   |   |   |--- NUM_EVENT_TYPES <= 17.50
|   |   |   |   |   |--- MAX_CAUSE_EVENTS <= 2.50
|   |   |   |   |   |   |--- class: Poor
|   |   |   |   |   |--- MAX_CAUSE_EVENTS >  2.50
|   |   |   |   |   |   |--- class: Fair
|   |   |   |   |--- NUM_EVENT_TYPES >  17.50
|   |   |   |   |   |--- MAX_CAUSE_EVENTS <= 2.50
|   |   |   |   |   |   |--- class: Fair
|   |   |   |   |   |--- MAX_CAUSE_EVENTS >  2.50
|   |   |   |   |   |   |--- class: Poor
|   |   |   |--- NUM_EVENT_TYPES >  22.50
|   |   |   |   |--- class: Poor
|   |   |--- MAX_INTERVENING_EVENTS >  0.50
|   |   |   |--- class: Poor
```

Figure 6 – Attention text decision tree.

The attention network primarily classifies instances by the number of cause events: a large block of instances fall into the *Fair* class when there is a single cause event. The number of causations is the next significant discriminator: better when 8 or less.

## NN

Scores: *Good*=14, *Fair*=19, *Poor*=48. The results are shown in Figures 7 and 8.

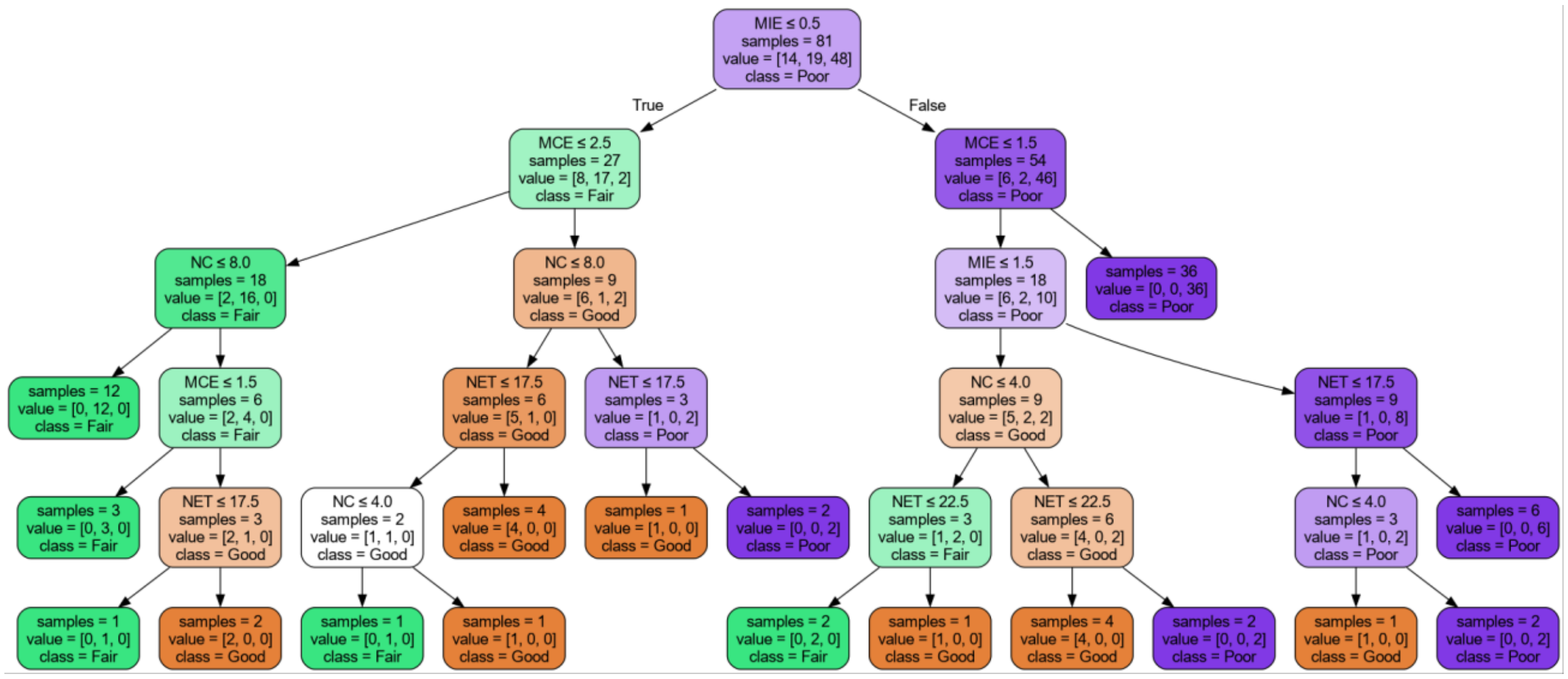

Figure 7 – NN graphical decision tree.

```
|--- MAX_INTERVENING_EVENTS <= 0.50
|   |--- MAX_CAUSE_EVENTS <= 2.50
|   |   |--- NUM_CAUSATIONS <= 8.00
|   |   |   |--- class: Good
|   |   |--- NUM_CAUSATIONS >  8.00
|   |   |   |--- MAX_CAUSE_EVENTS <= 1.50
|   |   |   |   |--- class: Good
|   |   |   |--- MAX_CAUSE_EVENTS >  1.50
|   |   |   |   |--- NUM_EVENT_TYPES <= 17.50
|   |   |   |   |   |--- class: Good
|   |   |   |   |--- NUM_EVENT_TYPES >  17.50
|   |   |   |   |   |--- class: Fair
|   |--- MAX_CAUSE_EVENTS >  2.50
|   |   |--- NUM_CAUSATIONS <= 8.00
|   |   |   |--- NUM_EVENT_TYPES <= 17.50
|   |   |   |   |--- NUM_CAUSATIONS <= 4.00
|   |   |   |   |   |--- class: Good
|   |   |   |   |--- NUM_CAUSATIONS >  4.00
|   |   |   |   |   |--- class: Fair
|   |   |   |--- NUM_EVENT_TYPES >  17.50
|   |   |   |   |--- class: Fair
|   |   |--- NUM_CAUSATIONS >  8.00
|   |   |   |--- NUM_EVENT_TYPES <= 17.50
|   |   |   |   |--- class: Fair
|   |   |   |--- NUM_EVENT_TYPES >  17.50
|   |   |   |   |--- class: Poor
|--- MAX_INTERVENING_EVENTS >  0.50
|   |--- MAX_CAUSE_EVENTS <= 1.50
|   |   |--- MAX_INTERVENING_EVENTS <= 1.50
|   |   |   |--- NUM_CAUSATIONS <= 4.00
```

```
|   |   |   |   |--- NUM_EVENT_TYPES <= 22.50
|   |   |   |   |   |--- class: Good
|   |   |   |   |--- NUM_EVENT_TYPES >  22.50
|   |   |   |   |   |--- class: Fair
|   |   |   |--- NUM_CAUSATIONS >  4.00
|   |   |   |   |--- NUM_EVENT_TYPES <= 22.50
|   |   |   |   |   |--- class: Fair
|   |   |   |   |--- NUM_EVENT_TYPES >  22.50
|   |   |   |   |   |--- class: Poor
|   |   |--- MAX_INTERVENING_EVENTS >  1.50
|   |   |   |--- NUM_EVENT_TYPES <= 17.50
|   |   |   |   |--- NUM_CAUSATIONS <= 4.00
|   |   |   |   |   |--- class: Fair
|   |   |   |   |--- NUM_CAUSATIONS >  4.00
|   |   |   |   |   |--- class: Poor
|   |   |   |--- NUM_EVENT_TYPES >  17.50
|   |   |   |   |--- class: Poor
|   |--- MAX_CAUSE_EVENTS >  1.50
|   |   |--- class: Poor
```

Figure 8 – NN text decision tree.

The NN (non-recurrent) performance is poor compared to the LSTM and Attention networks. The lack of intervening events is the most significant discriminator.

### Histogram

Scores: *Good*=81, *Fair*=0, *Poor*=0. The histogram algorithm performs very well, although some accuracies fall into the 90%+ range.

## Conclusion

The performance of the three investigated artificial neural network (ANN) architectures (LSTM, Attention, and non-recurrent) on tasks requiring the learning and classification of causations consisting of disordered cause events with intervening non-causal events was found to be generally fair. A conventional histogram algorithm significantly outperforms them, leading to a supposition that a hybrid system would be worth examination for tasks that require causation learning as a component. For example, a conventional algorithm would identify causations, passing them into a recurrent ANN, such as an LSTM or Mona network, which would process sequences of them.